\documentclass{article}
\usepackage{iclr2027_conference,times}

\usepackage{amsmath,amssymb}
\usepackage{graphicx}
\usepackage{booktabs}
\usepackage{hyperref}
\usepackage{url}
\hypersetup{hidelinks, pdftitle={Lingtai: What Concept Geometry Reveals---and Does Not Reveal---About LLM Inference}, pdfauthor={Jiangang Chen}}

\title{Lingtai: What Concept Geometry Reveals---and Does Not Reveal---About LLM Inference}

\author{Jiangang Chen\thanks{Chengdu Beiluoshimen Technology Co., Ltd.}}

\iclrfinalcopy

\begin{document}

\maketitle
\lhead{Under review as a conference paper at ICLR 2027}

\begin{abstract}
Observing what a large language model computes during autoregressive inference---online and without training probes---remains difficult. We introduce Lingtai, a training-free \textbf{concept telemetry} layer: at each generation step, residual states are projected onto a domain-specific bank of named concept anchors, constructed without labeled concept examples, outcome labels, gradient fitting, or activation-space optimization, producing a structured per-step concept-coordinate signal. Across code generation and grade-school mathematical reasoning, this signal exhibits a robust association with predictive uncertainty: the association survives problem-identity and token-position controls and is not attributable to a single token type, is not explained by a simple correct/incorrect mixture on GSM8K, and is not reproduced by matched random anchors; it is markedly weaker or direction-inconsistent in K-means and PCA projections. Two structures emerge: a recurring \textbf{uncertainty-linked activity} signal whose functional geometry is task-conditioned (distinct activity--entropy shapes on HumanEval, MBPP, and GSM8K), and an \textbf{execution-specific} trajectory identity with strong local inertia but weak re-instantiation invariance---under completion-only elastic alignment, corruption at $k=32$ (approximately a median quarter of the completion) on the matched re-execution subset still retrieves the archived episode at 62.0\%, while a fresh execution retrieves it only 11.7--16.0\% of the time. Finally, a matched audit finds \textbf{no evidence} that the scalar concept-activity signal used here supplies a stable correctness coordinate under the tested protocol; we therefore treat correctness as externally supplied. Telemetry adds 0.7--1.6\% per-token decode overhead for the 161-anchor code implementation, with unchanged generated tokens.
\end{abstract}

\section{Introduction}

Language-model inference is consumed as an opaque service: a prompt goes in, tokens come out, and the computation in between is invisible. Yet when the model is uncertain, how stable \emph{this} execution is, and whether a failure can be recognized again are questions about that hidden computation. Existing tools approach it from two sides: learned probes read internal states but need training data and can supply the structure themselves \citep{alain2017understanding,belinkov2022probing,hewitt2019designing}, and dictionary learning recovers sparse features at substantial training cost \citep{bricken2023monosemanticity,templeton2024scaling,huben2024sparse}; cheap output-level uncertainty measures \citep{guo2017calibration,kadavath2022language,kuhn2023semantic} do not directly expose the intermediate computation that produced the distribution. What remains less explored is an online measurement space with named coordinates and no concept or outcome labels---together with an explicit characterization of what it can and cannot reveal.

We introduce \textbf{Lingtai}, a training-free \emph{concept telemetry} layer. At each generation step, the residual stream at a chosen layer is projected through an RBF kernel onto a domain-specific bank of named concept anchors, yielding a concept-coordinate activity vector; scalar statistics of this vector (activity variance $K_{\mathrm{var}}$, effective participation) summarize a step. Anchors are constructed without labeled concept examples, outcome labels, or activation-space fitting---a co-occurrence network over domain terminology (Co-occurrence Network Concept Extraction, CNCE) yields the discovered concepts, and the math bank additionally appends 55 fixed numeric/operator primitives whose removal leaves the GSM8K result unchanged (Appendix~\ref{app:hardening}); embedding centers are extracted once per model. The same telemetry interface is instantiated in both domains (161 code and 1552 math anchors). We use ``high/low-activity'' only as operational quantile regions; the activity distribution is continuous, not bimodal.

We then ask what this signal \emph{reveals}. Across four benchmark--model configurations---HumanEval and MBPP with Qwen2.5-Coder-7B-Instruct, HumanEval with Qwen2.5-Coder-14B, and GSM8K with Phi-2---problem-level concept activity yields a positive paired log-ratio advantage over random anchors in 47/50, 47/50, 40/50, and 50/50 problems, with bootstrap CIs excluding zero. The association with predictive uncertainty survives problem-identity and token-position controls and is not attributable to a single token type; it is not explained by a simple correct/incorrect mixture in this GSM8K sample; and it is not reproduced by matched random anchors, while K-means/PCA retain far less of it with direction-inconsistent results. A diagnostic comparison shows internal contrast without external validity: PCA reaches far larger internal separation with weak or negative external correlation, while concept-activity contrast is modest but consistently positive.

Two robust but sharply bounded structures emerge---uncertainty-linked activity and execution-specific trajectory identity. We also ask what they do \emph{not} reveal. First, the association replicates but its functional form does not: HumanEval is monotone, MBPP U-shaped, GSM8K slow-then-steep. Second, execution identity has a sharp boundary: within a fixed execution the fingerprint is locally robust (corrupting a median quarter of the tokens still leaves the episode top-1 in 74--77\% of cases; exact replay gives cosine 0.99996), yet weakly invariant to re-instantiation---a fresh execution retrieves the same prompt's archived execution only $\sim$8\% of the time (10.6--21.3\% raw at the half-prefix), a paraphrase $\sim$3\%, and unrelated completions sit at the null. The fingerprint is thus a property of the particular execution path, not a stable semantic hash of the prompt; consistent identifier renames behave like arbitrary replacements, an absence of dissociation we treat as part of the result. Third, we test whether scalar concept activity could serve as a correctness coordinate. Under a position-matched audit, a pre-output signal that looks discriminative in naive analyses loses its advantage over simultaneously available legal baselines and shows no growth along generation. We find no evidence for a stable correctness coordinate from scalar concept activity under this protocol; we therefore treat correctness as externally supplied.

Telemetry costs 0.7--1.6\% per-token decode overhead for the 161-anchor code implementation, with unchanged generated tokens and $\sim$1\,MB extra peak GPU memory. We position the contribution as \textbf{measurement plus boundary}: a training-free signal that exposes task-conditioned activity structure and locally robust execution fingerprints, and controlled experiments that delimit what it does not reveal.

\noindent\textbf{Contributions.}
\begin{enumerate}
\item \textbf{Training-free structural telemetry.} We introduce a lightweight representation projecting residual states onto a bank of named concept anchors, exposing online activity structure at 0.7--1.6\% per-token decode overhead for the 161-anchor code implementation, with no change to generated tokens. \emph{(interface: code + math)}
\item \textbf{Two runtime phenomena.} Across code generation and grade-school math, concept activity carries uncertainty-related structure that survives problem-identity, token-position, token-type, and matched-baseline controls, and on GSM8K persists within correct and incorrect subsets---yet its functional form is task-dependent, and trajectory identity is execution-scoped: locally robust to corruption within a fixed execution, but brittle to re-instantiation. \emph{(coupling: code + math; fingerprint: code)}
\item \textbf{Boundary characterization.} A pre-specified support analysis finds common-mode-dominated activity, a diagnostic shows internal contrast without external validity, and under a position-matched audit we find no evidence for a stable correctness coordinate from the scalar activity statistic under that protocol. \emph{(support and contrast diagnostics: code + math; matched audit: tested code protocol)}
\end{enumerate}

\section{Related Work}

\noindent\textbf{Interpreting internal representations.} Linear probes established that intermediate states linearly encode task properties \citep{alain2017understanding,belinkov2022probing}, at the cost of per-property training and the risk that the probe supplies the structure---a concern made explicit by control-task designs \citep{hewitt2019designing}. Sparse autoencoders (SAEs) learn expressive sparse feature dictionaries from large activation corpora, recovering features that can be sparse, interpretable, and causally relevant \citep{huben2024sparse,bricken2023monosemanticity,templeton2024scaling}, but require a separate model- and layer-specific fitting stage. We treat SAEs as a learned-feature alternative rather than a dimension-matched baseline: a fair comparison would require training SAEs for our models. Lingtai targets a different operating point: domain-specific named coordinates with no activation-space optimization, intended as a runtime measurement layer rather than a sparse decomposition of the residual stream. Causal-intervention methods localize and edit specific circuits or facts \citep{meng2022locating,wang2023interpretability}. Concept telemetry is training-free, named by construction, and a measurement layer rather than an intervention; its anchors are domain-specific rather than universal---a trade we analyze rather than hide.

\noindent\textbf{Concept-based explanations.} Concept-based methods map representations to human-interpretable directions learned from labeled concept examples: TCAV \citep{kim2018tcav} tests sensitivity to a concept direction, FastCAV \citep{schmalwasser2025fastcav} reduces construction cost, and representation engineering \citep{zou2023representation} monitors population-level representations. Lingtai differs: anchors are constructed without labeled concept examples, and concept coordinates serve as an online telemetry space rather than an attribution or control mechanism.

\noindent\textbf{Uncertainty from outputs.} Output-distribution uncertainty measures are the standard cheap signals: calibration \citep{guo2017calibration}, self-evaluation of knowledge \citep{kadavath2022language}, and semantic-entropy variants that aggregate over meanings \citep{kuhn2023semantic}. These operate on outputs; telemetry reads the residual stream and overlaps predictive entropy only partially ($r = 0.518$ on HumanEval/Qwen2.5-Coder-7B-Instruct).

\noindent\textbf{Execution identity and fingerprints.} Prior work studies memorization and extraction from outputs \citep{carlini2021extracting}, and representation-similarity indices compare networks globally \citep{kornblith2019similarity}. Our fingerprint is neither a model identifier nor a semantic embedding: it is a per-episode trajectory signature locally robust under corruption but weakly invariant to re-instantiation (Section~\ref{sec:results-fingerprints}).

\noindent\textbf{Correctness signals from internal states.} A growing line of work reads hidden states for correctness and hallucination signals. CLUE \citep{liang2026clue} builds success/failure centroids from labeled experience and verifies new traces by centroid proximity---training-free, but requiring outcome labels; earlier work estimates hallucination risk from query-time internal states before response generation \citep{ji2024llm}, and other methods detect hallucinations from internal-state statistics \citep{chen2024inside,su2024unsupervised}. Recent work also exploits hidden representations for efficient uncertainty or difficulty estimation without repeated output sampling \citep{zhu2025llm,bu2026samplingfree}. These approaches produce verifiers or risk estimators; ours is complementary---a named-coordinate measurement layer without concept or outcome labels plus a controlled characterization of where activity and identity signals do and do not support correctness (Section~\ref{sec:results-gate2}). We make no claim to be first to read hidden states for correctness risk. Operationally, such detectors optimize risk estimation on intrinsic state statistics, whereas Lingtai fixes a domain-specific named frame with uncertainty as an external criterion only; we do not claim corpus anchors outperform intrinsic statistics at that task (raw hidden variance tracks entropy more strongly; Table~\ref{tab:audit}).

\section{Method}
\label{sec:method}

\subsection{Concept anchors (training-free, domain-specific)}
\label{sec:anchors}

For each domain we construct a bank of $K$ named concepts from a target-domain corpus, without concept labels, outcome labels, gradient-based fitting, or activation-space optimization:
(i) high-frequency domain terminology is extracted from the corpus;
(ii) a TF-IDF-weighted co-occurrence network over the terms is expanded by triangle detection and subgraph expansion into named concepts;
(iii) each concept's title is encoded by the target model itself, and the residual state at the title's final token (last-token residual) at the configuration's frozen extraction layer defines the anchor center $c_i$. We read the anchor bank as a fixed, dense, non-orthogonal measurement frame rather than a sparse concept disentangler: raw activity is common-mode dominated (Section~\ref{sec:results-commonmode}), and named coordinates are a human-readable reference system, not evidence of monosemantic features.
This yields \textbf{161 code} and \textbf{1552 math} anchors---the same telemetry interface instantiated in both domains (the math bank appends 55 fixed numeric/operator primitives to the discovered concepts; Appendix~\ref{app:hardening}). Anchor discovery is one-time, CPU-scale; at inference no anchors are learned or updated, and the bank is frozen before evaluation trajectories are extracted. Code anchors are discovered from a corpus of public Python code; math anchors from the GSM8K training split, while all GSM8K evaluations use held-out test problems. A decontamination audit over our 1,000-block anchor corpus against the HumanEval and MBPP problem, solution, and test text found no substantive overlap, and a parallel GSM8K train--test audit likewise found no substantive overlap (Appendix~\ref{app:hardening}).

\subsection{Concept telemetry}
\label{sec:telemetry}

\begin{figure}[t]
\centering
\includegraphics[width=0.74\linewidth]{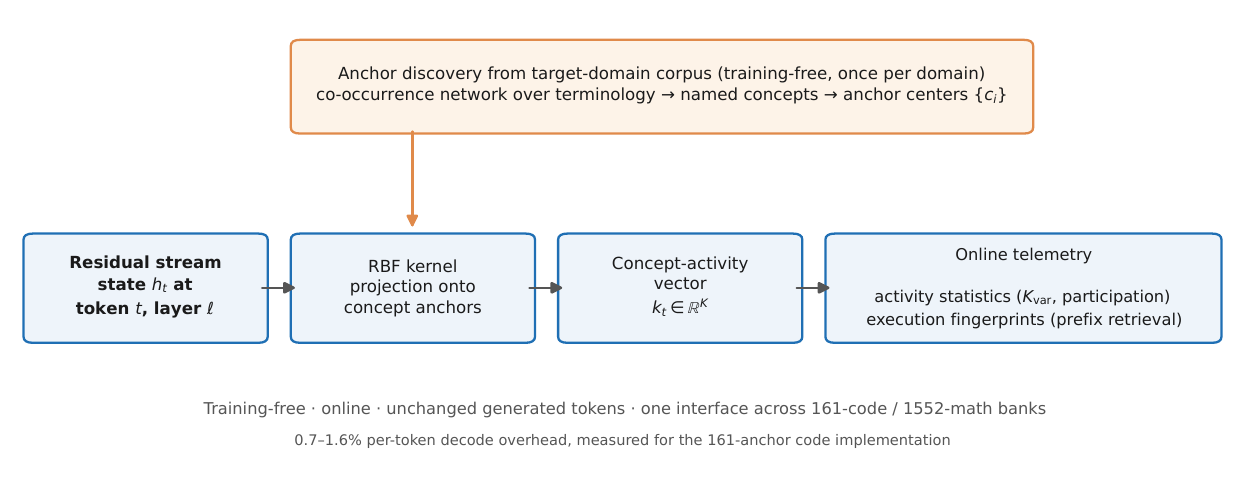}
\caption{Lingtai concept telemetry: residual stream at a chosen layer $\rightarrow$ RBF projection onto a domain-specific bank of named concept anchors $\rightarrow$ per-step concept-activity signal; anchors constructed without labeled concept examples or activation-space fitting; most are corpus-discovered, with fixed arithmetic primitives added only in the math bank.}
\label{fig:method}
\end{figure}

For any processed token position $t$, let $h_t$ be the residual state at a frozen layer. During online generation this is evaluated at each generated token; offline audits also process prompt positions (per-configuration analysis range: Appendix~\ref{app:hardening}). The telemetry vector (Figure~\ref{fig:method}) is the RBF similarity to the anchors,
\begin{equation}
k_t[i] = \exp\!\big(-\gamma\,(2 - 2\cos(h_t, c_i))\big), \qquad i = 1..K,
\end{equation}
with $\gamma = 1$. We report scalar statistics computed from $k_t$: activity variance $K_{\mathrm{var}} = \mathrm{Var}_i(k_t[i])$ across anchors, and effective participation $N_{\text{eff}} = 1/\sum_i p_i^2$ with $p_i = k_t[i]/\sum_j k_t[j]$. For episode retrieval, a trajectory is its sequence of telemetry vectors, similarity is the cosine over pooled prefixes, and the self-margin is computed against a frozen episode bank (Section~\ref{sec:scope}). The primary metric is \textbf{centered} (frozen-bank mean $\mu$ subtracted before pooling), because raw cosines are saturated by a shared common component; the raw metric is a secondary sensitivity throughout.

\subsection{Audit protocol for external validity}
\label{sec:audit-protocol}

All external-validity audits use the problem as the inferential unit (50 per configuration); pooled token-level statistics are descriptive. For each configuration we compute within-problem deciles of concept activity and compare mean predictive entropy across deciles; we test the association against matched controls: problem identity (within-problem deciles, per-problem paired statistics), token position (double demeaning and stratified analyses), token type (five lexical classes), and, on GSM8K, outcome stratification (correct vs incorrect generated trajectories). Matched baselines are random anchor banks, K-means clusters, and PCA projections at matching dimensions where applicable (scalar definitions and the AUROC target: Appendix~\ref{app:hardening}). The external criterion (predictive entropy) is never a function of the telemetry space.

\noindent\textbf{Pre-specified analyses.} Primary analyses and exit outcomes were fixed in advance: the raw-support analysis had two branches (Section~\ref{sec:results-commonmode}), the centered-profile analysis is \emph{exploratory}, the scope experiment's panels, perturbation classes, and monotonicity criterion were fixed (Section~\ref{sec:results-fingerprints}), and the matched audit follows the protocol specified after the naive analysis was invalidated (Section~\ref{sec:results-gate2}).

\subsection{Execution-fingerprint protocol}
\label{sec:scope}

We freeze a bank of 402 completed episodes (prompt + generated completion) from MBPP, with telemetry trajectories extracted once and never updated. For a query trajectory we compute the centered self-margin: the similarity to its own bank entry minus the maximum similarity to any other entry; top-1 denotes retrieval of the correct episode, and ``self'' always denotes the archived execution of the same task.

\begin{itemize}\itemsep0pt
\item \textbf{Panel A --- controlled local perturbation of a fixed execution.} We replace $k$ completion tokens by arbitrary pool tokens at positions sampled inside one of four bands (0.1/0.3/0.5/0.7 of the completion), re-encode by teacher forcing, and measure the self-margin (median over four bands at each $k$). A separate \textbf{identifier-renaming} condition replaces identifier tokens consistently (token-level, no retokenization) by same-class identifiers.
\item \textbf{Panel B --- natural re-execution.} Against the same frozen bank we ask whether a fresh stochastic re-execution of a prompt retrieves the archived execution of that same task as top-1: fresh samples (temperature 0.7; seeds 42 and 7; 94 bank tasks) and paraphrases of the same problem (seeds 42 and 7; 30 tasks), against a permutation null. Archived trajectories are greedy; a deterministic replay would reproduce its bank entry by construction, while Panel B instead samples a new execution.
\end{itemize}

\subsection{Cost}
\label{sec:cost}

We measure prefill (TTFT), per-token decode (TPOT), and throughput for a baseline versus a telemetry hook computing only the last-token projection and $K_{\mathrm{var}}$, on the same prompts, with warmup and three repetitions, verifying token-for-token identity between conditions.

\subsection{Configurations}
\label{sec:configs}

Four configurations are audited (50 problems each, one generation per problem; Appendix Table~\ref{tab:configs}): HumanEval \citep{chen2021evaluating} and MBPP \citep{austin2021program} with Qwen2.5-Coder-7B-Instruct \citep{hui2024qwen2} (NF4, layer 23); HumanEval with Qwen2.5-Coder-14B (INT8, layer 40); and GSM8K \citep{cobbe2021training} with Phi-2 \citep{javaheripi2023phi2} (FP16, layer 28). Every configuration's telemetry trajectory is extracted by teacher-forcing the recorded greedy completion---for GSM8K, Phi-2's own stored completion---so no model is re-sampled. Fingerprints use OpenCoder-1.5B-Base \citep{huang2025opencoder} (layer 10, inherited from the pre-existing scope-experiment protocol; 402-episode bank); the matched audit replays recorded runs with an independent re-generation check (layer 23, within the band where uncertainty coupling is strongest, Table~\ref{tab:layers}; 163 problems). Training-free means no gradient-based fitting, learned probe, outcome-supervised optimization, or labeled concept examples; anchors are one-time corpus preprocessing over GSM8K training-split problems (Section~\ref{sec:anchors}; statements with solutions; no test problem, correctness outcome, or generated completion). For the math bank, 1{,}497 anchors are corpus-discovered from the GSM8K training split and 55 fixed numeric/operator primitives are appended before center extraction (Appendix~\ref{app:hardening}).

\section{Results}

\subsection{External validity against predictive entropy, with matched baselines}
\label{sec:results-external}

\noindent\textbf{Problem-level association.} CNCE yields a positive paired log-ratio advantage over random anchors in 47/50 problems for HumanEval/Qwen2.5-Coder-7B-Instruct (mean $d_{\log} = 0.865$, 95\% CI [0.706, 1.025], Wilcoxon $p = 5.5\times10^{-13}$), 47/50 for MBPP/Qwen2.5-Coder-7B-Instruct ($d_{\log} = 0.629$, CI [0.497, 0.766], $p = 2.2\times10^{-12}$), 40/50 for HumanEval/Qwen2.5-Coder-14B ($d_{\log} = 0.674$, CI [0.471, 0.876], $p = 3.4\times10^{-8}$), and 50/50 for GSM8K/Phi-2 ($d_{\log} = 1.685$, CI [1.603, 1.766], $p = 1.8\times10^{-15}$; definitions: Appendix~\ref{app:hardening}). Token-level Spearman $\rho$ across the same four configurations is 0.297/0.236/0.235/0.443. Across ten independently drawn random banks, the positive count remains 42--50/50 (mean $d_{\log} = 0.811$, range [0.479, 1.360]); CIs are paired problem-level bootstraps (10{,}000 resamples) and $p$ values Wilcoxon signed-rank tests over the 50 paired problems.

\noindent\textbf{Matched baselines.} Table~\ref{tab:audit} reports the full HumanEval/Qwen2.5-Coder-7B-Instruct audit; direction replication is in Appendix~\ref{app:replication}.

\begin{table}[t]
\centering
\small
\setlength{\tabcolsep}{1.5pt}
\caption{External-validity audit, HumanEval / Qwen2.5-Coder-7B-Instruct (50 problems); criterion: predictive entropy; rank-based metrics. K-means and PCA are fitted transductively (in-sample, favorable to the baselines); K-means uses $k$-means++ ($n_{\mathrm{init}} = 10$) on all evaluated tokens. AUROC uses the top entropy decile over pooled tokens as the positive class.}
\label{tab:audit}
\begin{tabular}{lrccccr}
\toprule
Representation & Dim & Training & Named coordinates & Token $\rho$ & AUROC & Mean problem $\rho$ [CI] \\
\midrule
\textbf{Concept telemetry (ours)} & 161 & none & \textbf{yes} & 0.297 & 0.722 & 0.284 [0.238, 0.329] \\
Raw hidden & 3584 & none & no & 0.384 & 0.736 & 0.387 [0.356, 0.417] \\
K-means & 161 & unsup. & no & 0.102 & 0.630 & 0.140 [0.085, 0.192] \\
PCA top-10 & 10 & unsup. & no & $-0.070$ & 0.551 & 0.026 [$-0.028$, 0.073] \\
PCA top-20 & 20 & unsup. & no & $-0.015$ & 0.573 & 0.078 [0.031, 0.122] \\
PCA top-161 & 161 & unsup. & no & 0.033 & 0.570 & 0.118 [0.071, 0.164] \\
Random anchors & 161 & none & no & 0.011 & 0.499 & 0.004 [$-0.021$, 0.033] \\
\bottomrule
\end{tabular}
\end{table}

Telemetry (161 dimensions, no training, named coordinates) retains a problem-level $\rho$ of 0.284 [0.238, 0.329]; the raw residual stream (3584 dimensions) is higher, while K-means and the PCA projections reach far less. PCA baselines use the coordinate-variance scalar; a sign- and rotation-invariant projection-energy variant yields the same picture (Appendix~\ref{app:hardening}).

\noindent\textbf{Controls.} Within-problem deciles are monotone for HumanEval (d9/d0 = 7.1$\times$); the association survives double demeaning (residual Spearman: 0.23/0.20/0.20/0.42 across the four configurations). Token-type stratification identifies no single explaining class (all positive on HumanEval and GSM8K; small MBPP negatives). It is not explained by a simple correct/incorrect mixture in this GSM8K sample: stratifying by outcome yields 24/24 and 26/26 positive problems with near-identical advantages ($d_{\log} = 1.674$ vs 1.695; token-level Spearman 0.435 vs 0.460). Finally, $K_{\mathrm{var}}$ and predictive entropy are related but only partially overlapping (Pearson $r = 0.518$; Jaccard 0.422 on within-problem median-split token sets). The association is not an artifact of an under-tuned baseline, a single anchor draw, bandwidth, token-frequency confound, or prompt pooling: a completion-only sensitivity preserves the direction in all four configurations and the MBPP U-shape, and the five-layer sweep keeps L23 among the strongest layers (L26 slightly higher; Appendix~\ref{app:hardening}).

\noindent\textbf{Internal contrast is not external validity.} A diagnostic sweep shows internal separation and external association can diverge sharply: the largest PCA internal contrasts (up to 70.7$\times$) coincide with near-zero or negative external association, and PCA external validity is inconsistent across configurations (near-zero on HumanEval-14B, moderate on MBPP, negative at low rank on GSM8K); on MBPP, for example, a K-means representation reaches 2.3$\times$ internal contrast while its external $\rho = -0.10$, whereas telemetry's modest 2.2--5.2$\times$ contrast carries consistently positive association (0.22--0.44).

\subsection{The activity--entropy relation is task-conditioned}
\label{sec:results-geometry}

\begin{figure}[t]
\centering
\includegraphics[width=0.74\linewidth]{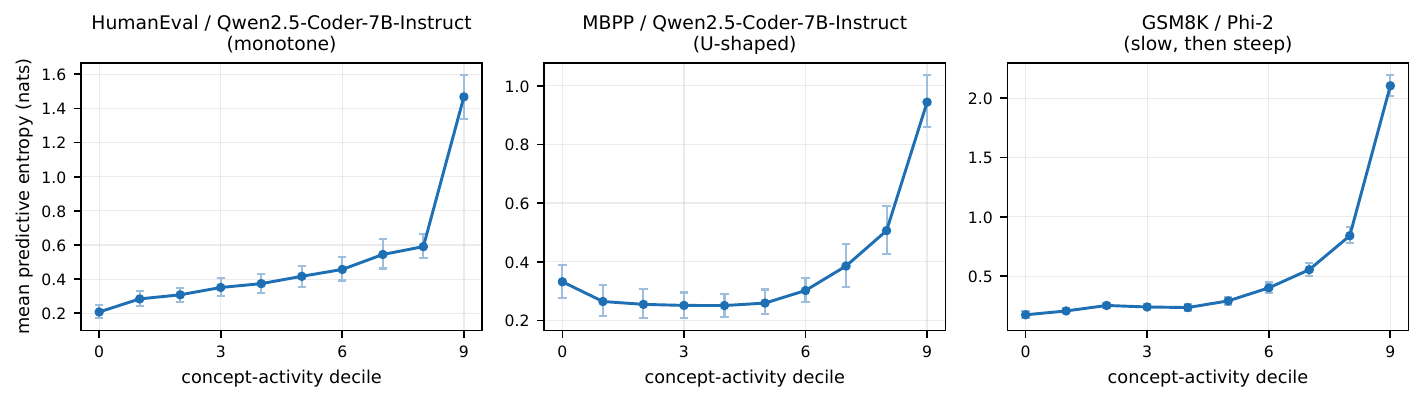}
\caption{The activity--entropy relation is task-conditioned, not universal. Within-problem deciles of mean predictive entropy by concept-activity decile (50 problems/panel; error bars = bootstrap 95\% CI).}
\label{fig:geometry}
\end{figure}

Within-problem decile curves (Figure~\ref{fig:geometry}): HumanEval is monotone (d9/d0 = 7.1$\times$), MBPP U-shaped (2.8$\times$), GSM8K slow-then-steep (12.0$\times$): the association replicates in direction, but its functional form does not. We therefore describe a \textbf{recurring activity--uncertainty coupling}, not a universal law; within the matched code configuration (same model and bank), HumanEval and MBPP already differ in form, and the GSM8K contrast is cross-domain and qualitative (different model and bank).

\subsection{What the activity signal consists of: common-mode and centered structure}
\label{sec:results-commonmode}

To understand what drives the activity--uncertainty coupling, we decompose the raw anchor profile into common-mode amplitude and centered variation: the raw profile turns out to be \textbf{common-mode dominated}. Effective participation saturates at the full dimension in both domains (MBPP: 160.1/160.4 of 161; math: 1526.4/1535.7 of 1552, high- vs low-entropy tokens), and community-level occupancy is likewise saturated: raw absolute activity is not a sparse concept selector.

As an algebraic separation $k_t = a_t\cdot\mathbf{1} + \delta k_t$ (amplitude plus centered profile), the empirical content is which component carries the variation. The decomposition highlights two components. (i) The amplitude itself is higher at high-entropy tokens (+0.027, +0.048, +0.039 across the three benchmarks), consistent with the common-mode interpretation of the raw signal. (ii) The \textbf{dispersion} of the deviation profile, $\mathrm{std}(\delta k_t)$, is systematically larger at high-entropy tokens: 0.0203 vs 0.0142 (MBPP), 0.0229 vs 0.0138 (HumanEval), 0.0244 vs 0.0152 (GSM8K), with the high side larger in \textbf{50/50 problems in each analysis} (paired CIs exclude zero), surviving position matching (both halves within a common token-position interquartile window). This dispersion is algebraically $\sqrt{K_{\mathrm{var}}}$---a re-expression of the primary finding rather than a second effect. The participation direction is task-dependent: the centered-profile participation ratio $\mathrm{PR}(\delta k_t^2)$ (element-wise square; Appendix~\ref{app:hardening}) rises at high entropy in GSM8K (49/50 problems) but in only 42\%/26\% of problems in the two code configurations. We label this analysis exploratory (Section~\ref{sec:limitations}).

\subsection{Execution fingerprints: locally robust, re-instantiation brittle}
\label{sec:results-fingerprints}

\begin{figure}[t]
\centering
\includegraphics[width=0.74\linewidth]{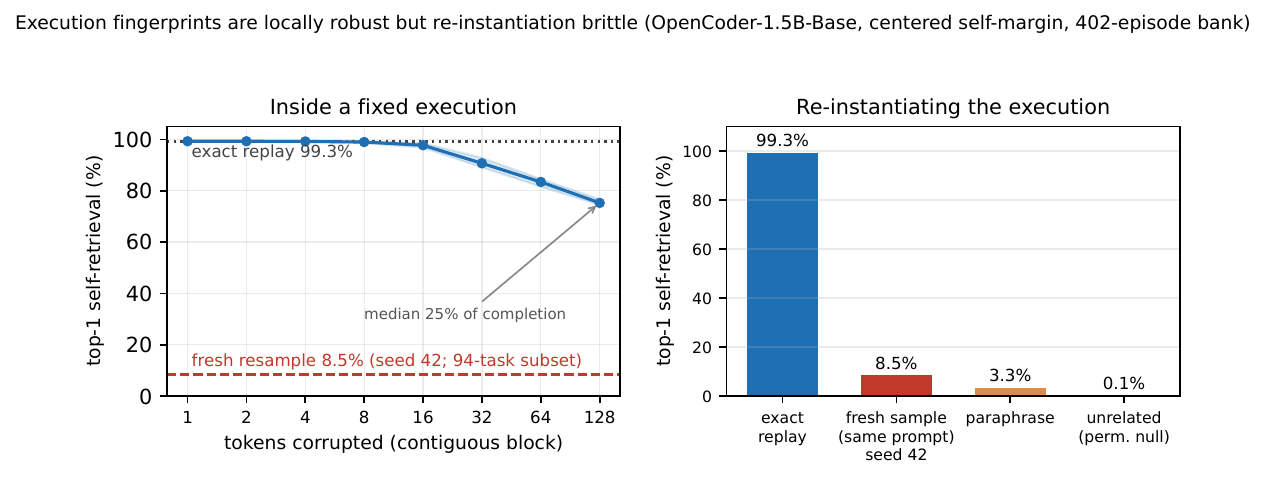}
\caption{Execution fingerprints are locally robust but re-instantiation brittle (OpenCoder-1.5B-Base, centered self-margin, 402-episode bank). Left: top-1 self-retrieval under a contiguous corruption block ($k = 1..128$; shaded range across four position bands; $k=128$ replaces a median 25\% of completion tokens). Right: exact replay vs fresh sample, paraphrase, and permutation null. Fresh-sample queries are the bank's 94 assertion/runtime episodes, re-sampled at temperature 0.7 under seeds 42 and 7 (left panel: 402-episode bank; dashed line: fresh re-execution rate on the 94-task subset, seed 42; right panel bars: seed 42); paraphrase queries are 30 tasks per seed.}
\label{fig:scope}
\end{figure}

\noindent\textbf{Exact replay.} Exact replay reproduces the fingerprint essentially perfectly (prefix cosine 0.99996; top-1 99.3\%).

\noindent\textbf{Controlled local perturbation (Panel A, Figure~\ref{fig:scope}, left).} Corrupting a contiguous block of completion tokens degrades the fingerprint gracefully and monotonically: mean self-margin (centered, full prefix) falls from 0.375 at $k=1$ to 0.083--0.095 at $k=128$, with top-1 self-retrieval from 99.3\% to 74--77\%. The decline is strictly monotone in each band (Spearman $\rho=-1.000$); position effects are negligible at small $k$ and mildly negative at $k=32$ ($\rho=-0.117$). At $k=128$, the corruption replaces a \textbf{median 25\% of completion tokens} (short completions force window expansion; capped mean 38.5\%). A text-cosine baseline stays at $\Delta \le 0.104$ (Appendix~\ref{app:hardening}).

\noindent\textbf{Surface renames are not special.} Consistent identifier renames ($m = 1, 2, 4$) give $\Delta s = 0.371/0.361/0.351$ with top-1 $\sim$99\%, the same trend as arbitrary replacements; this \emph{absence of dissociation} is consistent with execution-path continuity and shows no evidence of special invariance to consistent identifier renaming.

\noindent\textbf{Natural re-execution (Panel B, Figure~\ref{fig:scope}, right).} Against the same frozen bank, a fresh sample from the same prompt retrieves the archived execution of that task in 8.5\% of cases (centered, full prefix; seed 42; 7.4\% for seed 7), a paraphrase in 3.3\%, and unrelated queries sit at the permutation null (0.1\%). The two regimes differ sharply on a matched task set: restricting Panel~A to the same 94 episodes, the grid point nearest a median quarter of these completions ($k=32$) retains top-1 self-retrieval in 60--75\% versus 29.8--30.9\% at $k=128$, while a fresh execution of the same tasks succeeds in only 8.5\%/7.4\%. Execution identity is tied to the particular realization of a run, not to a stable semantic identity of the prompt: Panel~B re-executes the identical prompt, so prompt-prefix identity cannot explain the local-robustness vs.\ re-instantiation contrast. An alignment stress test shows the contrast is not a rigid token-index artifact: alignment-tolerant rescoring does not erase the gap---under completion-only rescoring, fresh re-execution retrieval stays at 11.7--16.0\% (DTW) and 10.6--12.8\% (lag-max), against 62.0\% and 46.3\% for $k=32$ corruption on the matched re-execution subset (Appendix~\ref{app:hardening}).

\subsection{No stable correctness coordinate from scalar concept activity under the matched protocol}
\label{sec:results-gate2}

\begin{figure}[t]
\centering
\includegraphics[width=0.74\linewidth]{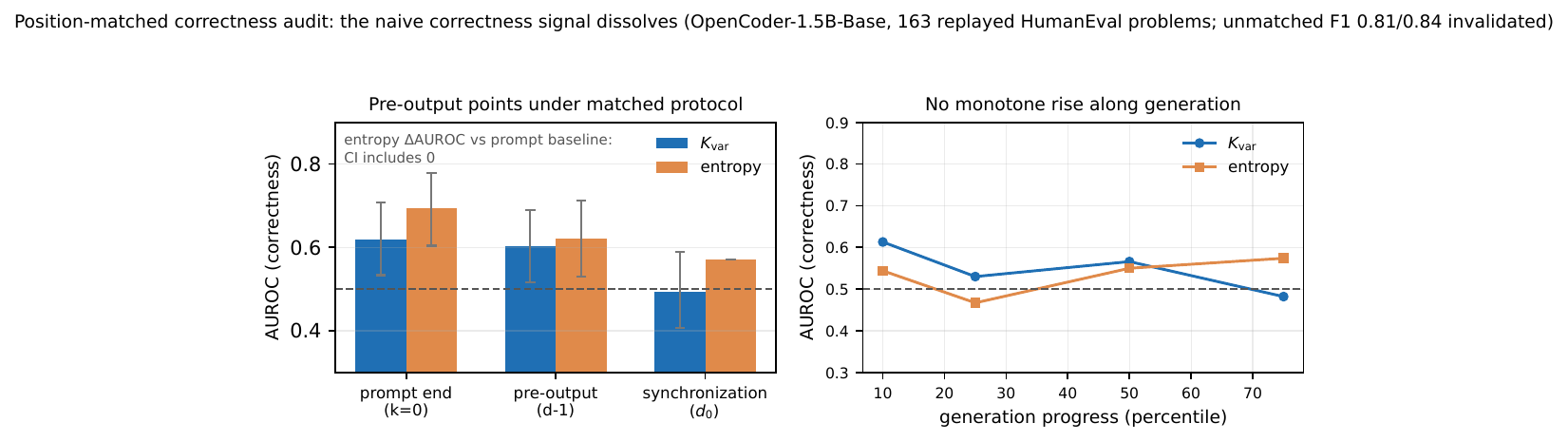}
\caption{Position-matched correctness audit (OpenCoder-1.5B-Base, 163 replayed HumanEval problems). Left: pre-output AUROC for $K_{\mathrm{var}}$ and predictive entropy at the prompt end ($k=0$), the pre-output divider ($d-1$), and the synchronization point ($d_0$); the paired $\Delta$AUROC for entropy against the prompt-average-entropy baseline has CIs including 0 at all three points. Right: no monotone rise along generation progress. The earlier unmatched F1 = 0.81/0.84 claim is invalidated.}
\label{fig:gate2}
\end{figure}

An earlier analysis suggested concept activity detects erroneous continuations before emission (naive F1 = 0.809/0.842; pooled divergence tokens vs.\ unmatched passing positions, tokens treated as independent); it does not survive the position-matched audit. The audit replays 163 of 164 HumanEval runs (83 failing, 80 passing; one excluded for replay fidelity below 0.8, mean 0.976); $d$ is each run's divergence step (first departure from the canonical solution; $d{=}0$ for 84\% of failing runs), with passing runs assigned pseudo-divergence positions sampled from the failing distribution (definitions: Appendix~\ref{app:hardening}). At the prompt end ($k=0$), AUROC for $K_{\mathrm{var}}$ is 0.619 [0.533, 0.709]; at the pre-output divider ($d-1$) 0.604 [0.517, 0.691]; at synchronization ($d_0$) it falls to 0.494 [0.408, 0.589]---chance. Entropy shows the same behavior (0.694/0.621/0.571). Against prompt-derived baselines (length 0.642, entropy 0.622), the paired $\Delta$AUROC for entropy has CIs containing zero at all three points, $K_{\mathrm{var}}$ never exceeds them, and AUROC does not rise monotonically (Figure~\ref{fig:gate2}). \textbf{We find no evidence that scalar concept activity supplies a stable correctness coordinate under this matched protocol}---the representation carries task-conditioned activity structure and locally stable, execution-scoped identity, but neither property constitutes a stable correctness signal under this protocol. We retain only the weaker statement that $K_{\mathrm{var}}$ partially overlaps with predictive entropy.

\subsection{Cost}
\label{sec:results-cost}

The hook adds 0.71\%/1.57\% per-token decode time on Qwen2.5-Coder-7B-Instruct (NF4) and OpenCoder-1.5B-Base (FP16), with token-identical outputs and $\sim$1\,MB extra peak memory (throughput reductions 0.72\%/1.54\%). Projection cost scales linearly in $K$; the math bank was not profiled (Section~\ref{sec:limitations}).

\section{Discussion}

\noindent\textbf{Two structural findings, sharply bounded.} Telemetry exposes two structures. First, an uncertainty-linked activity signal recurs across two domains and two model families, but its functional geometry and participation structure are task-conditioned (Sections~\ref{sec:results-geometry},~\ref{sec:results-commonmode}) and its coordinate basis is domain-specific. Second, execution trajectories exhibit strong local inertia but weak re-instantiation invariance, including under alignment-tolerant rescoring (Section~\ref{sec:results-fingerprints}). We therefore treat both as bounded empirical structures rather than universal properties of LLM inference.

\noindent\textbf{Two complementary properties, neither a stable correctness signal as operationalized here.} The telemetry has two faces---uncertainty-linked activity structure and an execution-scoped trajectory fingerprint---and local corruption tolerance far exceeds re-instantiation invariance.

\noindent\textbf{Observation is endogenous; correctness is treated as externally supplied.} Striking internal structure can coexist with no stable correctness advantage under a fair protocol (Section~\ref{sec:results-gate2}); we do not propose a detector: telemetry's value is measurement, not classification.

\noindent\textbf{Architectural implication: scope-bound incident memory.} Because correctness is externally supplied while execution identity is real, an external outcome (a test verdict, a human label) could be \emph{bound to the episode that produced it} and retrieved only when a similar episode recurs; Panel~B quantifies the scope ($\sim$8\% fresh, $\sim$3\% paraphrase). Such memory would retrieve a labeled failure trajectory without generalizing the mistake---an architectural implication, not a built system.

\section{Limitations}
\label{sec:limitations}

\begin{enumerate}\itemsep0pt
\item \textbf{Fingerprint evidence is code-only.} Fingerprints use only OpenCoder-1.5B-Base on MBPP-style episodes; Panel B's paraphrase set is small ($n=30$/seed), and trajectory similarity is index-aligned in the main analysis; an elastic-alignment stress test preserves the reported contrast (Appendix~\ref{app:hardening}), while cross-layer sweeps remain future work.
\item \textbf{The centered-dispersion analysis is exploratory.} Section~\ref{sec:results-commonmode} was not pre-specified, has no held-out confirmation, controls position but not token type, and its participation component is task-dependent; the raw-support analysis is negative.
\item \textbf{The correctness boundary is protocol-bound.} The matched audit covers one code model and 163 replayed problems; it neither supports a correctness coordinate beyond this protocol nor rules out correctness information under others, and it tests only scalar $K_{\mathrm{var}}$---directional or subspace structure was not tested.
\item \textbf{Domain-specific banks; cost scope.} Anchors are domain-specific; cross-domain claims are directional replications, never numerical comparisons. Cost audits cover only the 161-anchor code bank (not the math bank); Panel A uses a single perturbation seed.
\end{enumerate}

\enlargethispage{3\baselineskip}
\section{Conclusion}

Lingtai exposes two structures: an uncertainty-linked activity signal and an execution-specific trajectory identity. The first recurs across tasks but is task-conditioned; the second has strong local inertia yet weak re-instantiation invariance. Neither is a general correctness signal: under the tested protocol, scalar concept activity supplies no stable correctness coordinate.

\noindent\textbf{AI use statement.} Generative AI tools supported ideation, literature search, design critique, research code, data analysis, figure/code edits, drafting, and paraphrase stimuli for a controlled audit. No formal proof relies solely on AI-generated reasoning; all AI-assisted code was validated against recorded artifacts. The authors take responsibility for the content. \textbf{Reproducibility statement.} All experiments use public benchmarks and open-weight models. The supplementary package contains protocols, frozen artifacts and checksums, configuration details, reproduction code, and claim-to-result mappings for the reported analyses. \textbf{Ethics statement.} This work uses only public benchmarks (HumanEval, MBPP, GSM8K) and involves no human subjects or personal data. An execution fingerprint recognizes a specific past generation and is locally robust to corruption yet brittle to re-instantiation; in deployment it would be computed from model-internal states available to the serving operator rather than to users. Our results give no evidence of cross-session identification; we nevertheless flag fingerprinting as privacy-relevant and release no data enabling cross-user tracking.

\clearpage
\bibliography{references}
\bibliographystyle{iclr2027_conference}

\appendix

\section{Replication details}
\label{app:replication}

\begin{table}[h]
\centering
\scriptsize
\setlength{\tabcolsep}{4pt}
\caption{Configurations and extraction protocol. For the four external-validity configurations, trajectory telemetry is extracted by teacher-forcing a recorded greedy completion; no model is re-sampled during extraction. Process selection: the 50-problem audit subsets are deterministic ID-ordered prefixes fixed before analysis, with no selection on telemetry or outcome (GSM8K: first 50 of the test split). Panel B fingerprint queries are separately re-sampled at temperature 0.7 (seeds 42/7); all other trajectories are greedy. Fingerprint row: 402-episode bank; 94 fresh-sample and 30 paraphrase queries per seed.}
\label{tab:configs}
\begin{tabular}{lllr}
\toprule
Benchmark & Model (precision), layer & $n$ & Trajectory source \\
\midrule
HumanEval, MBPP & Qwen2.5-Coder-7B-Instruct (NF4), L23 & 50 each & recorded greedy completion \\
HumanEval & Qwen2.5-Coder-14B (INT8), L40 & 50 & recorded greedy completion \\
GSM8K & Phi-2 (FP16), L28 & 50 & stored own completion (replay) \\
MBPP fingerprints & OpenCoder-1.5B-Base (FP16), L10 & 402; 94$\times$2; 30$\times$2 & Panel A teacher-forced; B re-sampled \\
Matched audit & OpenCoder-1.5B-Base (FP16), L23 & 163 & recorded + re-generation check \\
\bottomrule
\end{tabular}
\end{table}

Table~\ref{tab:replication} reports direction replication across configurations.

\begin{table}[h]
\centering
\small
\setlength{\tabcolsep}{5pt}
\caption{Direction replication of the concept-telemetry row: problems with a positive paired $d_{\log}$ against random anchors, and token-level Spearman $\rho$, per configuration. Configurations differ in model, quantization, layer, and anchor bank; we compare direction only, never absolute values.}
\label{tab:replication}
\begin{tabular}{lrr}
\toprule
Setting & Problems with positive paired $d_{\log}$ & Token $\rho$ \\
\midrule
HumanEval / Qwen2.5-Coder-7B-Instruct & 47/50 & 0.297 \\
MBPP / Qwen2.5-Coder-7B-Instruct & 47/50 & 0.236 \\
HumanEval / Qwen2.5-Coder-14B & 40/50 & 0.235 \\
GSM8K / Phi-2 & 50/50 & 0.443 \\
\bottomrule
\end{tabular}
\end{table}

For a step $t$, let $\omega_j = \delta k_t^2[j]/\|\delta k_t^2\|_1$; the centered-profile participation ratio is $\mathrm{PR}(\delta k_t^2) = 1/\sum_j \omega_j^2$, the centered-profile analogue of the effective number of active coordinates, and is reported as the per-problem mean within entropy halves; at high entropy it rises in GSM8K (49/50 problems) and in only 42\%/26\% of problems in the two code configurations (CI contains zero for MBPP, below zero for HumanEval). Position matching restricts both halves to the pooled interquartile range of within-problem token positions.

Natural re-execution (Panel B; Clopper--Pearson 95\% CI; centered full-prefix metric): top-1 retrieval is 8.5\% [3.8, 16.1] for a fresh sample (seed 42, $n=94$) and 7.4\% [3.1, 14.7] for seed 7, and 3.3\% [0.1, 17.2] for paraphrases (1/30 in each of seeds 42 and 7; $n=30$ per seed); under the raw half-prefix metric the fresh-sample rates are 21.3\% [13.5, 30.9] (seed 42) and 10.6\% [5.2, 18.7] (seed 7); the permutation null sits at 0.1\%.

\section{Hardening controls and definitions}
\label{app:hardening}

\noindent\textbf{Per-representation scalars (Table~\ref{tab:audit}).} Each representation yields one scalar per token.
(i) \emph{CNCE}: $K_{\mathrm{var}} = \mathrm{Var}_i(k_t[i])$ with $k_t$ as in Eq.~1 ($\gamma=1$).
(ii) \emph{Random anchors}: the same construction with 161 fixed Gaussian unit vectors.
(iii) \emph{K-means}: the same construction with 161 cluster centers as anchors; we fit $k$-means++ with $n_{\mathrm{init}}=10$ on all evaluated tokens (in-sample, favorable to the baseline): token $\rho$ 0.102, AUROC 0.630, mean problem $\rho$ 0.140 [0.085, 0.192], contrast 2.9$\times$; a single-initialization 2{,}000-token variant gives 0.083/0.614/0.116.
(iv) \emph{PCA top-$k$}: variance across the token's first $k$ centered principal-component projections (transductive, in-sample).
(v) \emph{Raw hidden}: variance across all 3{,}584 dimensions.

\noindent\textbf{Internal contrast.} For a scalar signal $s$, internal contrast is the ratio of its mean over the high half of tokens to its mean over the low half at the median split; per-problem reports split within problem, baseline sweeps split pooled tokens.
The external criterion is the predictive entropy $H_t = -\sum_v p_t(v)\log p_t(v)$ of the model's next-token distribution at token $t$, computed over all prompt$+$completion tokens (teacher-forced, $\le$512 tokens, no temperature scaling) for the code configurations, and over the final-question segment plus the stored completion (few-shot shared prefix excluded) for GSM8K. A completion-only sensitivity check (analysis restricted to completion-token positions; same tokenizer) preserves the association in all four configurations (positive token-level $\rho$; problem-level 95\% CIs excluding zero; the MBPP U-shape persists), with attenuated effect sizes (\texttt{completion\_only\_sensitivity.json}). The AUROC positive class is the top decile of entropy over the pooled evaluation tokens; AUROC 0.722 reads as the probability that a top-decile-entropy token has higher concept-activity variance than a lower-entropy token.

\noindent\textbf{Problem-level statistic.} For problem $p$, $R = \bar H_{\text{high}}/\bar H_{\text{low}}$, where the halves are the tokens at or above vs.\ below the problem's median $K_{\mathrm{var}}$; $d_{\log} = \log R_{\text{CNCE}} - \log R_{\text{random}}$ is paired by problem. The sampling unit is the problem: CIs are paired bootstraps over problems (10{,}000 resamples) and $p$ values Wilcoxon signed-rank tests over the 50 paired values.

\noindent\textbf{Divergence and replay definitions (Section~\ref{sec:results-gate2}).} For a failing run, $d$ is the first completion-token index at which the completion departs from the canonical solution under token-level sequence alignment; $d-1$ denotes the last pre-divergence position and $d_0$ the divergence position. Here $d$ is an alignment index, not a semantic error-onset annotation. Replay fidelity is the fraction of positions at which the teacher-forced greedy argmax equals the recorded completion token; one of 164 runs fell below the 0.8 threshold and was excluded (mean fidelity 0.976).

\noindent\textbf{Fingerprint similarity (exact form).} A trajectory is its sequence of telemetry vectors $k_1,\dots,k_T$ at the audit layer, and the frozen-bank mean subtracts the token-level mean over all bank tokens, $\mu = \big(\sum_b T_b\big)^{-1}\sum_b\sum_{j=1}^{T_b} k^{(b)}_j$, from every query and bank trajectory ($\bar k = k-\mu$). With $\langle\cdot,\cdot\rangle$ the flattened-prefix inner product and prefix length $\kappa = \min(\lceil pT_q\rceil, T_q, T_b)$, the similarity to bank entry $b$ is $s_p(q,b) = \langle \bar q_{1:\kappa}, \bar b_{1:\kappa}\rangle / (\|\bar q_{1:\kappa}\|\,\|\bar b_{1:\kappa}\|)$; the self-margin is $\Delta s_i = s_p(q,b_i) - \max_{j\neq i} s_p(q,b_j)$, and top-1 denotes $\Delta s_i > 0$. The primary metric is centered full prefix ($p=1$); the raw variant drops centering, and the half-prefix variant uses $p=0.5$. Corruption replaces a contiguous block of $k$ completion tokens sampled inside a position band centered at $10/30/50/70\%$ of the completion; when $k$ exceeds the band width the window expands, so for short completions the block covers the whole completion.

\noindent\textbf{Matched-subset check for the fingerprint contrast.} Panel~B's fresh re-executions cover the bank's 94 assertion/runtime episodes, fixed by that task-type criterion across the whole bank before any re-sampling (a strict superset of the earlier 30-task set; 158 newly generated trajectories plus 30 reused; see the supplementary artifact log), not selected by retrieval outcome. Their completions are shorter (median 108 tokens vs.\ 512 for the remaining 308); fixed-$k$ corruption is therefore harsher for them, and the panels are not directly comparable across task sets. Restricting Panel~A to those same 94 tasks (centered, full prefix), top-1 self-retrieval at $k=128$ is $29.8$--$30.9\%$ across position bands (Wilson 95\% CI for $28/94$: $[21,40]\%$), and at $k=32$, the grid point nearest a median quarter of these completions ($\approx 27$ tokens), it is $60$--$75\%$ (bands $0.1/0.3/0.5/0.7$: $74.5/59.6/60.6/69.1\%$), against $8.5\%$/$7.4\%$ for a fresh execution of the same tasks (seeds 42/7) and $88.0$--$90.9\%$ for the remaining 308 tasks at $k=128$. The qualitative contrast is therefore unchanged on a matched task set.

\noindent\textbf{Alignment stress test (Section~\ref{sec:results-fingerprints}).} To test whether the fingerprint contrast is an artifact of rigid token-index alignment, all four conditions are rescored with two alignment-tolerant metrics on completion-only trajectories: (i) \emph{lag-max}, the maximum centered cosine over global shifts $\ell \in [-16,16]$ of the overlapping window, and (ii) \emph{DTW}, the normalized elastic path cost with local cost $1-\cos$ and a Sakoe--Chiba band of $|\Delta T| + 10\%$ of the longer trajectory; every one of the 402 bank candidates is scored under the identical metric, centered by the frozen bank mean. Top-1 self-retrieval is: exact replay $1.000$/$1.000$, Panel-A $k=32$ (mean of four bands) $0.463$/$0.620$, fresh re-execution $0.128$/$0.160$ (seed 42) and $0.106$/$0.117$ (seed 7) for lag-max/DTW, against an empirical permutation null of $\approx 0.1\%$ (uniform 402-way chance $\approx 0.25\%$). Alignment-tolerant rescoring therefore does not erase the gap---fresh re-execution retrieval stays at 11.7--16.0\% under DTW and 10.6--12.8\% under lag-max, against 62.0\% and 46.3\% for $k=32$ corruption---so the local-inertia vs re-instantiation asymmetry is not a rigid-index artifact (\texttt{alignment\_stress\_audit.json}).

\noindent\textbf{Random-bank seeds.} Across ten random banks (seeds 42--51) the positive-problem count stays at 42--50/50 (mean $d_{\log} = 0.811$, range [0.479, 1.360]); token-level $\rho$ is near zero on average (mean 0.032, range [$-0.162$, 0.147]).

\begin{table}[h]
\centering
\small
\caption{Kernel-bandwidth sensitivity (HumanEval / Qwen2.5-Coder-7B-Instruct).}
\label{tab:gamma}
\begin{tabular}{lccc}
\toprule
$\gamma$ & Token $\rho$ & Mean problem $\rho$ & Mean $d_{\log}$ \\
\midrule
0.25 & 0.295 & 0.282 & 0.874 \\
0.5 & 0.297 & 0.283 & 0.867 \\
1.0 & 0.297 & 0.284 & 0.865 \\
2.0 & 0.297 & 0.284 & 0.855 \\
4.0 & 0.296 & 0.284 & 0.844 \\
\bottomrule
\end{tabular}
\end{table}

\noindent\textbf{Kernel bandwidth.} Varying $\gamma$ over $\{0.25, 0.5, 1, 2, 4\}$ (Table~\ref{tab:gamma}) leaves the association essentially unchanged; $\gamma=1$ is not a special point.

\begin{table}[h]
\centering
\small
\caption{Layer sweep with per-layer anchor centers (last-token title residuals re-encoded at each layer).}
\label{tab:layers}
\begin{tabular}{lcc}
\toprule
Layer & Token $\rho$ & Mean problem $\rho$ [CI] \\
\midrule
5 & $-0.110$ & $-0.084$ [$-0.114$, $-0.053$] \\
13 & 0.115 & 0.131 [0.102, 0.160] \\
20 & 0.262 & 0.249 [0.206, 0.286] \\
23 & 0.297 & 0.284 [0.238, 0.328] \\
26 & 0.306 & 0.307 [0.268, 0.347] \\
\bottomrule
\end{tabular}
\end{table}

\noindent\textbf{Layer sweep.} With anchor titles re-encoded at each layer, the association grows with depth and is broadly positive from mid layers onward (Table~\ref{tab:layers}); layer 23 is not a cherry-picked outlier (layer 26 is marginally stronger; the L23 entry is re-estimated in this sweep, so its bootstrap CI endpoint differs from Table~\ref{tab:audit} in the third decimal).

\noindent\textbf{Token-frequency control.} Rank-based partial correlations of $K_{\mathrm{var}}$ and entropy controlling token log-frequency (estimated from the 1{,}000-block code corpus, 336{,}316 tokens), position, and problem leave the association intact: raw $\rho = 0.297$; controlling log-frequency 0.298; adding position 0.323; adding problem 0.323; the per-problem mean partial $\rho$ controlling frequency and position is 0.275 [0.233, 0.319]. Token type enters the main-text stratification.

\noindent\textbf{PCA sign-invariance audit.} The coordinate-variance PCA scalar is not invariant to arbitrary PCA axis signs; recomputing every configuration with a sign- and rotation-invariant projection-energy variant ($\sum_j z_j^2$) changes token $\rho$ by at most 0.06 and AUROC by at most 0.05, leaves low-rank/negative PCA external validity unchanged, and a scale-free capture-ratio variant is weaker still; the weak external validity observed for PCA is therefore not an artifact of axis-sign ambiguity (\texttt{pca\_invariance\_audit\_*.json}).

\noindent\textbf{Anchor-corpus decontamination.} We checked all 1{,}000 corpus blocks against 2{,}656 HumanEval/MBPP text fields (prompt, entry point, canonical solution, tests) using normalized 8-gram and exact-line matching. Five blocks were flagged for review---three exceeding 10\% 8-gram coverage and two matching exact generic lines; manual review of the longest shared subsequences found only generic loop or test-literal sequences (e.g., a 14-token nested-loop skeleton), with no substantive overlap. The same audit for the 1{,}552 math anchors against the GSM8K test split finds no exact-line matches, one block above 10\% 8-gram coverage (11\% on a generic ``miles/hours'' phrase), and verbatim multi-word titles only for generic mathematical vocabulary or arithmetic expressions (e.g., ``half of'', ``20 / 4''), consistent with shared problem templates rather than substantive leakage (\texttt{decontam\_math\_anchors.json}).

\noindent\textbf{CNCE parameters and title encoding.} For the math bank, co-occurrence edges join terms whose TF-IDF cosine exceeds 0.1 (at most 10 neighbors per term; the builder retries at 0.05/15 when too few concepts emerge), and triangle detection over this graph seeds subgraph expansion (density $\ge$0.05, size 2--8) into multi-word titles. Titles are encoded bare (no chat template), and the anchor center is the last non-padding token's residual at the extraction layer. Full constants and tie-breaking rules are fixed in the released builders (\texttt{build\_gsm8k\_concept\_network.py}, \texttt{build\_concept\_anchors.py}).

\begin{table}[h]
\centering
\scriptsize
\setlength{\tabcolsep}{3pt}
\caption{Five code and five math concepts (co-occurrence degree, source snippet, nearest anchors with cosine).}
\label{tab:anchor-examples}
\begin{tabular}{llp{8.2cm}}
\toprule
Concept & Deg. & Source snippet $\rightarrow$ nearest anchors \\
\midrule
typer app name & 100 & \texttt{import typer; app = typer.Typer()} $\rightarrow$ order data list (0.69); import license module (0.68) \\
int float list & 78 & \texttt{class FocalNetConfig(...)} $\rightarrow$ order data list (0.83); node none list (0.82) \\
parser args add\_argument & 38 & \texttt{parser = argparse.ArgumentParser()} $\rightarrow$ args parser add\_argument (0.98) \\
license import lightning & 32 & \texttt{\# Copyright 2025 the HuggingFace Team} $\rightarrow$ torch import license (0.54) \\
tensor self torch & 32 & \texttt{def \_quantize\_tensor(...)} $\rightarrow$ model self torch (0.94) \\
00 hours minutes & 8 & ``Josh has soccer practice on Monday\ldots'' $\rightarrow$ hours hour minutes (0.83) \\
000 100 20 & 8 & ``MegaCorp got caught leaking\ldots'' $\rightarrow$ 500 10 50 (0.85) \\
10 erasers boxes & 8 & ``Mary bought 14 apples, 9 oranges\ldots'' $\rightarrow$ tissues boxes packs (0.60) \\
100 10 110 & 8 & ``Diane is playing poker\ldots'' $\rightarrow$ bills 10 100 (0.73) \\
1000 customers bought & 8 & ``A watermelon stand sold 46 watermelons\ldots'' $\rightarrow$ profit 50 sells (0.71) \\
\bottomrule
\end{tabular}
\end{table}

\noindent\textbf{Anchor examples.} Table~\ref{tab:anchor-examples} shows five code and five math concepts with their co-occurrence degree, a source snippet, and nearest anchors. We present these as named coordinates, not as evidence of human concept-faithfulness.

\begin{table}[h]
\centering
\scriptsize
\setlength{\tabcolsep}{3pt}
\caption{Positioning relative to learned-feature alternatives.}
\label{tab:positioning}
\begin{tabular}{lcccc}
\toprule
& Learned probe & SAE & TCAV/CAV & Lingtai \\
\midrule
Outcome/property labels & usually yes & no & concept exemplars & \textbf{no} \\
Activation-space fitting & yes & yes & yes & \textbf{no} \\
Model/layer-specific learned representation & yes & yes & yes & anchors re-embedded \\
Named coordinates & property-specific & post-hoc & yes & \textbf{by construction} \\
Goal & prediction & sparse decomposition & attribution & \textbf{runtime telemetry} \\
\bottomrule
\end{tabular}
\end{table}

\noindent\textbf{Learned-feature alternatives.} Table~\ref{tab:positioning} places Lingtai next to probes, SAEs, and concept-vector methods. We do not run a numerical SAE comparison: a fair one would require training SAEs for our evaluated models or changing the base model to one with public pretrained SAEs, and the two approaches target different questions.

\noindent\textbf{One-time preprocessing and determinism.} Concept discovery is CPU-dominated; a clean rebuild of the math bank takes roughly 10--21\,min on our machine (624\,s and 665\,s for the deterministic builder under two hash seeds, with the co-occurrence network and concept birth dominating in those runs; an earlier build took 1{,}249\,s). Tie-breaking in the released builder is deterministic (stable lexical keys, explicit secondary keys, deterministic concept identifiers): two clean rebuilds under different hash seeds produce byte-identical artifacts (1{,}643 total anchors, including the same 55 fixed primitives; checksums in the supplementary material). A deterministic rebuild is comparable to but not identical to the frozen 1,552-anchor bank used here (tie-resolution order differs); all audits use the frozen artifacts, released with SHA256 checksums. Exact numerical reproduction of the reported results uses the released frozen 1{,}552-anchor bank and its checksum; the revised deterministic builder is provided for reproducible future reconstruction. For the frozen math bank, 55 fixed numeric/operator primitives (digits, operators, and generic computation patterns) are appended to the discovered concepts before center extraction, so 1{,}497 of the 1{,}552 anchors are discovery-derived (the 161 code anchors contain no such additions); excluding these 55 primitives leaves the reported GSM8K association essentially unchanged (token $\rho$ 0.443 vs.\ 0.440; paired $d_{\log}$ 1.685 vs.\ 1.677; 50/50 positive problems; \texttt{math\_anchor\_extras\_ablation.json}).

\noindent\textbf{Cost protocol.} Overhead was measured on a single RTX 3090 (batch 1, prompts $\le$512 tokens, 64 generated tokens, five HumanEval prompts), computing only the last-token RBF projection and $K_{\mathrm{var}}$, with three warmup and three timed repetitions per condition and token-for-token output comparison between baseline and hook; the 1{,}552-anchor math bank was not measured.

\end{document}